%% file: main.tex
\pdfoutput=1
\documentclass[11pt]{article}
\usepackage{authblk}
\usepackage[]{acl}

\usepackage{times}
\usepackage{latexsym}
\usepackage{tabularx}
\usepackage{graphicx}
\usepackage{float}
\usepackage{amsmath}
    \newcolumntype{L}{>{\raggedright\arraybackslash}X}

\usepackage[T1]{fontenc}
\usepackage[utf8]{inputenc}

\usepackage{microtype}

\usepackage{xcolor}

\title{Analogy Generation by Prompting Large Language Models:

A Case Study of InstructGPT}
\author[1]{Bhavya Bhavya}
\author[2]{Jinjun Xiong}
\author[1]{ChengXiang Zhai}
\affil[1]{Department of Computer Science, University of Illinois at Urbana-Champaign}
\affil[1]{\texttt{\{bhavya2, czhai\}@illinois.edu}}
\affil[2]{Department of Computer Science and Engineering, University at Buffalo}
\affil[2]{\texttt{jinjun@buffalo.edu}}

\begin{document}
\maketitle
\begin{abstract}
%Pre-trained Language Models (PLMs) such as BERT and GPT have been applied to many tasks of text generation (e.g., summarization, machine translation, dialogue system). 
We propose a novel application of prompting Pre-trained Language Models (PLMs) to generate analogies and study how to design effective prompts for two task settings: generating a source concept analogous to a given target concept (aka Analogous Concept Generation or \textsc{acg}), and generating an explanation of the similarity between a given pair of target concept and source concept (aka Analogous Explanation Generation or \textsc{aeg}).   
%Given a target concept, we study how to design prompts to prompt a PLM to generate an analogous source concept with an explanation of the similarity between the target and the source concepts. 
%We found that We also 
We found that it is feasible to prompt InstructGPT to generate meaningful analogies and the best prompts tend to be precise imperative statements especially with a low temperature setting. We also systematically analyzed the sensitivity of the InstructGPT model to prompt design, temperature, and injected spelling errors, and found that the model is particularly sensitive to certain variations (e.g., questions vs. imperative statements). Further, we conducted human evaluation on 1.4k of the generated analogies and found that the quality of generations varies substantially by model size. The largest InstructGPT model can achieve human-level performance at generating meaningful analogies for a given target while there is still room for improvement on the \textsc{aeg} task.\footnote{Our code and datasets are available for public use: https://github.com/Bhaavya/InstructGPT-Analogies} 
%Additionally, we study how to evaluate the generated analogies. For this, we
% We also investigated the suitability of using the existing reference-based metrics designed for evaluating natural language generation (NLG) to evaluate analogy generation and found that the recent BLEURT score is better than the others. We further propose a promising consensus measure
%based on consensus 
% based on diverse prompts and settings, which can be potentially
% used to both automatically evaluate the generated analogies in the absence of reference text (e.g., in novel domains) and rank a set of generated analogies to select analogies of different characteristics. 
%Generating analogies have a wide range of applications, such as explaining concepts and scientific innovation. 
% Overall, our study shows that InstructGPT offers a promising new way to generate natural language analogies of concepts, breaking the limitation of
% existing analogy generation methods in requiring structured representation. 

%Existing approaches to generation of analogies rely on structured representations and thus can only work on limited domains. Our study shows
% and that with appropriate prompting, PLMs can not only generate useful analogies mentioned in the training data used for training the PLMs, but also generate novel analogies not (directly) mentioned in the training data. Furthermore, PLMs can also be prompted to generate personalized analogies that can effectively explain a concept to each individual using a potentially different analogous concept adapted to the background of the individual.   
\end{abstract}

\input{intro}
\input{rel_work}
\input{method}

\input{experiments}

\section{Limitations}

A major limitation of our study is that we only studied analogies on a small reference dataset in
%usefulness of PLMs for generating analogies in only 
one domain (high-school science).  Our newly created reference data sets are relatively small due to limited resources found online. But, the sample size of the automatically generated analogies we evaluated was large ($\sim$ 31k automatically evaluated, and $\sim$ 1.4k manually evaluated) thereby mitigating some concerns about bias due to small dataset size. Moreover, as our research questions study an open-ended generation task, having a pre-defined list of reference candidates is not ideal for evaluation. Thus, future research is required to more thoroughly evaluate the generated analogies and investigate  the generalizability of the findings to other domains. 

Further, the manual evaluation was conducted by a selected group of people in the US and might not reflect the opinions of a more diverse group. Moreover, our kappa scores of 0.3-0.5, although common in previous NLG evaluation work \cite{van2021human}, are not on the higher end. In general, thresholds to determine what counts as high or low kappa scores tend to be open to interpretation \cite{van2021human}. Thus, we've released our annotated and full datasets online, as also suggested in \cite{van2021human}, and invite other researchers to further investigate them.

\vspace{-0.2cm}
\section{Conclusion}
\vspace{-0.2cm}
In this study, we proposed and studied the novel task of generating analogies by prompting InstructGPT. Our experiments showed that the InstructGPT is effective on this task when precise prompts are used, thus offering a promising new way to generate analogies, which can break the limitation of the traditional analogy generation methods in requiring a pre-generated structured representation.   

By evaluating the performances of the various designed prompts in multiple temperature settings and in the presence of synthetic spelling errors, we found that the InstructGPT model is sensitive to those variations (e.g., question vs. imperative-style prompts). Additionally, based on human evaluation, we found that the quality of the generated analogies substantially depends on the model size. The largest model was found to achieve human-level performance at generating analogies for given target concepts. There is still much room for improvement at the challenging task of explaining the analogical similarity between the given target and source concepts. 

Our work opens up many exciting opportunities for future work both for application-oriented and foundational research on PLMs for analogy generation. For example, conducting more robustness analyses based on prompt perturbations (e.g., natural spelling mistakes, grammar, length, etc.). Also, in addition to the unsupervised approaches we explored in this paper, it is interesting to develop supervised approaches for this task including by fine-tuning PLMs on our created datasets.
% Additionally, we studied the automatic evaluation of generated analogies and confirmed that the recent BLEURT metric is more effective compared to others. We also designed a promising, diversity-based consensus score for evaluation. 

% Our work opens up many exciting opportunities for future research including developing methods for analogy generation in multiple domains and metrics for automatic evaluation of generated analogies, and intelligently designing effective prompts and choosing temperature settings of PLMs.  
\section{Acknowledgments}
This work is supported in part by the IBM-Illinois Center for Cognitive Computing Systems Research (C3SR) as an IBM AI Horizon’s Network.

\section{Ethical Considerations}
The risks associated with using PLMs for analogy generation are similar to those of NLG tasks, such as bias, toxicity, and misinformation \cite{bender2021dangers,weidinger2021ethical}. Accordingly, these should be carefully evaluated before deploying the models for any practical applications, such as education. 

Furthermore, there is a steep monetary and environmental cost associated with using the GPT-3 models, especially Davinci. The OpenAI API charges \$0.06 /1K tokens. Including early experiments, analogy generation in this study costed a total of about \$240. Since we conducted multiple runs with the same prompt account for randomness (e.g., in the high temperature setting), the costs rose sharply. 
% Future work should investigate the capabilities of other smaller and more cost-effective models for this task. 
%  \jx{Maybe add some details to explain where the cost is coming from}.

\bibliography{ref}
\bibliographystyle{acl_natbib}
\input{appendix}

\end{document}

%% file: intro.tex
\section{Introduction}

 \begin{table}[htbp]
  \vspace*{-\baselineskip} 
    \centering
     \caption{Selected prompts and InstructGPT-generated analogies for \textit{natural selection}}
 \begin{tabular}{p{2cm} p{5cm}}
    \hline
\textbf{Prompt} (P7): & \textit{What is analogous to natural selection?} \\
 \textbf{InstructGPT Output}: & The analogous process to natural selection is artificial selection.    \textbf{(9 words)} \\
    \hline
\textbf{Prompt} (P2): & \textit{Explain natural selection using a well-known analogy.} \\
\textbf{InstructGPT Output}: & Imagine that you have a jar of mixed nuts ... If you shake the jar ...the big nuts will fall out first ... analogy is that natural selection is like a sieve that separates the fit from the unfit... \textbf{(136 words)} \\
    \hline
  \label{tab_sample_std}
  \end{tabular}
  \vspace*{-15mm}.
    \end{table}
   
Large Pre-trained Language Models (PLMs) such as BERT\cite{devlin2018bert} and GPT\cite{brown2020language} have been applied to many tasks of text generation (e.g., summarization, dialogue system) with promising results~\cite{li2021pretrained}. However, no existing work has studied how to apply PLMs to generate different kinds of textual analogies, such as conceptual metaphors (e.g.,``Life is a journey\footnote{https://en.wikipedia.org/wiki/Conceptual\_metaphor}''), and instructional analogies (e.g., ``A red blood cell is like a truck in that they both transport essential supplies''\cite{newby1995instructional}).

Generating analogies has a wide range of applications, such as explaining concepts and scientific innovation, and analogies play a crucial role in human cognition. Analogical matching and reasoning enables humans to understand and learn unfamiliar concepts (aka target concepts) by means of familiar ones (aka source concepts) and to make scientific innovations.
% e.g. the Wright brothers developed a steerable aircraft based on bicycles. 
Unsurprisingly, analogy modeling and generation has been a long-standing goal of AI \cite{mitchell2021abstraction}. This is a challenging problem because it often requires computing deep semantic similarities that are beyond the surface-level similarity. For example, the Bohr's atom model and the solar system are analogous due to their structural and relational similarities (i.e., atoms orbit around the nucleus like planets around the sun).

Much work has been done to compute such analogical similarities between concepts. However, existing approaches mostly rely on structured representations, thus, they can only where such representations already exist. For example, one of the most popular models is Structural Mapping Engine (SME) \cite{forbus2017extending}, which aligns \textit{structured representations} of the target and source concepts using predicate logic. Moreover, they cannot \textit{generate} analogies in natural language. 
% \jx{The example didn't explain the issue on ``limited domains''.} 
% To get such representations more easily, one potential approach is to automatically extract them. However, current automatic extraction of structured representations, e.g. using Semantic Parsing of text, is complex and error-prone \cite{ribeiro2021combining}. Thus, it is more natural to use other approaches that can be directly applied to unstructured text.

Inspired by the recent success in applying PLMs to many NLP tasks (e.g., ~\cite{li2021pretrained}), we propose and study the application of PLMs to analogy generation. We consider two typical application scenarios of analogy generation: 1) Analogous Concept Generation (\textsc{acg}): given a target concept (e.,g, bohr's model), generate a source concept analogous to the target concept (e.g., solar system), possibly with an explanation of their similarities;
2) Analogy Explanation Generation (\textsc{aeg}): given a target concept and an analogous source concept, generate an explanation of their similarities. 

%We note that both tasks defined above involve the generation of text segment $Y$ given another text segment $X$ as input. 
%conditional text generation tasks
%study how to design effective prompts to prompt a PLM to generate a source concept analogous to a given target concept as well as to generate an explanation of the similarity between given pair of target concept and source concept.

%We define the task of analogy generation as the following: for any given input text $X$ that describes a target concept, generate another text $Y$ that describes a source concept analogous to the target concept, possibly with some explanation of the similarity of $Y$ and $X$. For example, an analogy for Bohr's model could be ``Bohr's model is like a solar system. The nucleus is the sun and the electrons are planets orbiting around it.''

By noting the similarity of the two tasks defined above to 
other text generation problems, and being inspired by the recent success of using prompted PLMs for text generation, we propose analogy generation by using a PLM with appropriately designed prompts. We adopt the promising emerging paradigm of prompting language models \cite{liu2021pre} that uses textual prompts with unfilled slots and directly leverages the language models to fill those slots and obtain the desired output. For example, Table \ref{tab_sample_std} shows sample prompts and PLM-generated outputs for \textsc{acg} from our experiments.

% Turney et al. \cite{turney2008latent} studied such analogies, however, only for automatically \textit{choosing} the mappings between source and target concepts and attributes, which is a highly restricted setting. The problem of automatically \textit{generating} such analogies has not been studied in any previous work.  
% {\em {\color{red} Old text: Recently, language modeling has been successful for a wide variety of natural language tasks. For example, Word2vec \cite{mikolov2013efficient}, BERT \cite{devlin2018bert}, GPT-3 \cite{brown2020language} perform well on simpler, word-level analogical reasoning (e.g., king is to man as queen is to ?), although their feasibility for more complex analogical matching (e.g. Bohr's model is analogical to ?) has not been studied before. }}

% Large pre-trained language models (e.g., GPT-3) have already been successful for other tasks like question answering with minimal further training for downstream tasks. 
% Along this direction, P

% Firstly, analogical reasoning and explanation require deep knowledge of the attributes, functions and relations of both source and target concepts.  Secondly, analogies are most useful when they explain target concepts using everyday-life scenarios. Both commonsense reasoning and creativity are required to generate plausible yet interesting analogical texts (e.g., electrical resistance is like cats blocking mice). It is unclear whether PLMs are capable of such tasks. 
% Moreover, since this is a new problem, it is unclear how to evaluate the quality of the generated analogies.

Specifically, we 
% we tackle those challenges
%explore the potential of using PLMs for analogy generation 
study the following main research questions: RQ1) How effective is a modern PLM such as InstructGPT in generating meaningful analogies? RQ2) How sensitive are the generated analogies to prompt design, the temperature hyperparameter, and spelling errors? RQ3) How does the model size impact the quality of generated analogies?

% RQ2) Are existing NLG evaluation metrics suitable for evaluating generated analogies and can they give meaningful results?  RQ3) How sensitive are the generated analogies to prompt design and other hyperparameters? RQ4) Can we automatically assess analogies in the absence of reference dataset? 

% RQ1) How effective is a modern PLM such as GTP-3 in generating meaningful analogies? RQ2) To what extent does the effectiveness of PLMs for analogy generation depend on the discussion of a target concept in the training data used for training the PLM (i.e., the Web)? Can a PLM generate novel analogies that were not explicitly mentioned on the Web? RQ3) How can we prompt a PLM to generate analogies that are adapted to the background of a user for explaining a concept? 

To study these questions, we design several experiments on analogies generated from the InstructGPT \cite{ouyang2022training} model. First, we manually validate whether InstructGPT can generate meaningful analogies for ten well-known analogies in the science domain. Next, we design and systematically vary prompt variants (e.g., imperative statements vs. questions) and temperature, and investigate the corresponding variations in the generated text by comparing them to a reference dataset of science analogies. Finally, we study the impact of model size on the quality of generated analogies both by automatically comparing against the reference data and using human evaluation.
% Firstly, to assess whether existing NLG metrics (e.g., BLEURT \cite{lin2004rouge}) can be reliably used for evaluating the quality of generated analogies, we design two sanity tests. The tests check whether the metrics generally behave as expected, i.e. give higher scores to higher quality analogies. 
%generated by better methods. 
% Using these tests, we select the best metric for automatically evaluating the InstructGPT generated analogies against a reference dataset of analogies of middle-school science concepts created from online academic Q\&A sites.  Finally, since it is not always feasible to obtain reference datasets (e.g., for new domains or novel analogies), we design a scoring method based on consensus of generated analogies in various settings (e.g., prompt design, temperature), called Consensus Score, to automatically evaluate the analogies without reference text. We also investigate its effectiveness as an evaluation method based on its correlation with BLEURT. 

Our experimental results show that PLMs (specifically, InstructGPT) offer a promising general approach to generating analogies with properly designed prompts.
%using prompts. 
Furthermore, the InstructGPT model is found to be sensitive to the prompt design, temperature, and spelling errors for this task, particularly to the prompt style (i.e., question vs. imperative statement). Precise imperative statements in low-temperature setting are found to be the best prompts. Finally, the quality of the generated analogies depends heavily on the model size. While the largest model can achieve human-level performance on the \textsc{acg} task, the smallest model barely generates any meaningful analogies. The \textsc{aeg} task proved to be more challenging based on human evaluation and could be a better test of the analogical reasoning capabilities of PLMs especially for explaining analogies not seen during training.

% We also confirm the effectiveness of the recent BLEURT\cite{sellam2020bleurt} score for evaluating analogies . We  show that our Consensus Score metric with diversity offers a promising way of reference-free evaluation of generated analogies. 
% in unrestricted domains and that with appropriate prompting, PLMs can not only generate useful analogies mentioned in the training data used for training the PLMs, but also generate novel analogies not (directly) mentioned in the training data. Furthermore, PLMs can also be prompted to generate personalized analogies that can effectively explain a concept to each individual using a potentially different analogous concept adapted to the background of the individual.   

%Inspired by the recent success of prompting, in this work, we propose a general framework for finding analogous source concepts for a given target concept based on prompting. For this, we will 
% compute the analogical similarities between source and target descriptions. Directly estimating the analogical similarity between them can require extensive training data \cite{hope2017accelerating}. It also also not interpretable. Thus, as an intermediate step, we will use prompting to extract analogical structure including attributes and relations of the base and target and mappings between them. This structure will be used to estimate the similarity. 

%% file: rel_work.tex
\section{Related Work}
\subsection{Computational Models of Analogies}
There has been a lot of work on computational modeling of analogies \cite{mitchell2021abstraction}. The SME model \cite{forbus2017extending} is one of the most popular symbolic model that finds the \textit{mapping}, or connections between structured representations of source and target concepts and their attributes. However, such methods cannot generate new analogous source concepts with analogical explanation. 
% Similarly, Turney et al. \cite{turney2008latent} studied the problem of automatically choosing all the analogical mappings between source and target concepts and attributes. 

% However, their discriminative task setting is not practically useful since it requires evaluating the similarities between every possible combination of concepts and their attributes. 
The recent deep learning-based approaches, including using pre-trained language models \cite{mikolov2013efficient, rossiello2019learning, ushio2021bert}, are able to \textit{generate} analogies to some extent, but are currently limited to simple word-level and proportional analogies, such as (ostrich:bird :: lion:?). In contrast, we aim to generate and explain more complex analogies of concepts, e.g. instructional analogies \cite{newby1995instructional}.

Another line of work is on finding analogous documents for scientific innovation, such as product descriptions and research papers, based on their semantic similarities \cite{kittur2019scaling}. In contrast, we operate in a generative task setup. 
% They learned vector-based representations for each aspect of a document (e.g. \textit{purpose}) and then computed vector similarities between documents. 

To the best of our knowledge, none of the existing work has studied the problem of automatically generating complex analogies in natural language. Recently, research on more ``generative'' analogy-making tasks has been recommended \cite{mitchell2021abstraction}. Along this direction, we believe that our proposed task is challenging and more practically useful than the existing text-based generative analogical tasks including letter-string (e.g., if ``abc'' changes ``abd''), what
does ``pqrs'' change to?) and word-level analogies. 

% \cz{I'm thinking of removing all the following: Indeed, in simple scenarios, such as proportional analogies
% (e.g., Paris: France:: London: ?), the
% analogical lexical relation reasoning can be supported  by language models ~\cite{ushio2021bert} to some extent. However, in those simple analogies, the source and target concepts are typically words that are similar to each other only along a single relation (e.g., captial-of). In this paper, we are interested in generating more challenging and elaborated analogies where the source and target concepts (or situations) could have several similarities that all should be explained in the generation of $Y$.  }
.      
% Analogy finding from text has also been studied in other applications, e.g., for instruction \cite{kumar2014automatically}. However, they extracted existing analogical descriptions and did not find new analogical relations.  
\vspace{-7mm}
\subsection{Prompting Language Models}

Recently, prompts have been either manually created or learned to successfully leverage PLMs for several natural language tasks \cite{liu2021pre}. Our work is closest to prompting for lexical and proportional analogy generation \cite{ushio2021bert}. But, none of the existing work has performed an in-depth study on prompting PLMs for both generating analogous concepts given a single query concept and explaining the analogical similarities between two query concepts.

% which has recently shown great promise  \cite{khashabi2020unifiedqa,tafjord2021general}. Analogy generation could be considered as a special and challenging case of question answering. For example, two questions in the Challenge300Dataset \cite{tafjord2021general} require comparing the relation between two entities (e.g., how do pandas and parrots differ). But, none of the existing work has performed an in-depth study on prompting PLMs for both generating analogous concepts given a query concept and explaining the analogical similarities between concepts.

% question answering and information extraction including extracting events, relation and entities from text \cite{li2019entity, du2020event, cui2021template}. They design or learn prompts to extract known facts, relations, entity types, roles, and arguments from large pre-trained language models. We are inspired by this line of work. Analogical structure extraction is special because the expected number or types of entities and relations is not specified and varies for each case. We will design suitable prompts for this task. 

%% file: method.tex
\section{Problem Formulation}

%Formally, our task is to estimate the conditional probability that text segment $Y$ provides a meaningful analogy explanation of a text segment containing a target concept $X$, $p(Y|X)$, and seek the output $Y^*$ that has the highest conditional probability, i.e., 
%$$Y^* = \arg\max_Y p(Y|X)$$. 

 %Overall, the task is to generate an analogy $Y$ given an input $X$ that maximizes the conditional probability, $P(Y|X)$. 
 
 Motivated by the practical applications of this task (e.g., explaining concepts), we study analogy generation in the following settings.
 
1. Analogous Concept Generation (\textsc{acg}) or \textbf{No Source (\textsc{no\_src})}: Here, only the target concept is provided as the input.
%i.e., $X$ contains a target concept for which we seek the analogy (e.g., atom). 
The goal is to generate an analogous source concept or scenario, along with some explanation to justify the analogy. For example, ``Explain Bohr's atomic model using an analogy.''
% Practically, this setting could be useful in finding unknown analogies and creating novel analogies. 

2. Analogy Explanation Generation (\textsc{aeg}) or \textbf{With Source (\textsc{wsrc})}: Here, in addition to the target, the source concept is also a part of input. The goal is to generate an explanation of how the target and source are analogous. For example, ``Explain how Bohr's atomic model is analogous to the solar system.''

Our problem setup is similar to the use of PLMs for text generation \cite{li2021pretrained}, and is most closely related to single-relation analogy generation (e.g., ostrich : bird :: animal : lion) \cite{ushio2021bert}, where the input is  a pair of query concept (e.g., ostrich : bird), and the task is to choose an analogical pair from a pre-defined list of candidate pairs. But, our proposed task is still  different in nature and much more challenging (e.g., requiring more creativity in some cases). First, both of our inputs and outputs are different. For example, in the proposed \textsc{acg} setup, our input is a single concept (e.g., ``bohr's model''), not a pair of concepts. Our task is to identify another concept (or scenario) that has an equivalence to the query concept based on their deep and non-trivial semantic similarities. No previous work has studied this kind of ``single-concept-based'' analogy generation with pre-trained language models. Even in the proposed \textsc{aeg} setup where we also use a pair of concepts as input, they are different from the pair used in the previous work.  For example, our input could be a pair (e.g., ``bohr's model'' and ``solar system'') and the output is an explanation of their analogical relations (e.g., how their structures are similar). Second,  we do not have a pre-defined finite list of candidates to choose from, which is a more realistic and interesting setting than previous work from application perspectives, and is also much more challenging for evaluation. 

% If successful, this setup could be useful in assisting users in creating better analogical explanations and seeing connections between two seemingly disparate concepts or situations.

% 3. \textit{Adaptive}: In this setup, in addition to the target, some background or preferences of the user are also a part of $X$. The expected $Y$ should contain an analogous an source concept along with explanation, which is adapted to the user.

%% file: experiments.tex
\section{Experiment Setup}

In this section, we discuss InstructGPT PLM and datasets used in our experiments. 

%\section{Experiments}
%In this section, we discuss the experiments performed to investigate our research questions. We first introduce the reference dataset we created in Section \ref{data}. \jx{missing section 4.2?} Then, we study the feasibility of PLMs for generating analogies, Section \ref{feasibility}. Next, we investigate the effectiveness of existing reference-based NLG metrics \ref{auto_eval} and select the best metric for further study. In Section \ref{sec_hypsen}, we discuss the impact of prompt designs and temperature settings on the generated analogies. Finally, in Section \ref{conscore}, we design a consensus-based score for reference-free automatic evaluation of generated analogies. 

%\subsection{Dataset}
%\label{data}

%\subsection{GPT-3 Davinci Instruct Model}
\noindent {\bf InstructGPT Model:}
Recently, several PLMs have been developed and trained on massive web data  \cite{devlin2018bert,brown2020language,raffel2019exploring}. In this study, we probe the aligned GPT-3 models, InstructGPT. These are GPT-3 models that have been optimized to follow instructions better \cite{ouyang2022training}. InstructGPT has four variants depending on the model size (number of parameters), namely Ada (350 M), Babbage (1.3 B), Curie (6.7 B) , and Davinci (175 B)\footnote{https://blog.eleuther.ai/gpt3-model-sizes/}. Unless otherwise mentioned, we use the Davinci model for the experiments as it is expected to have the best performance.

We used the Open AI API \footnote{https://beta.openai.com/docs/api-reference/completions/create} to generate all analogies. Main hyperparameters are described in Section \ref{temp} and rest in the Appendix \ref{hyperparam}
% \cz{provide a justification of why 939 was chosen as the parameter value.}

\noindent {\bf Dataset:}
As the task of analogy generation, as defined in this paper, has not been previously studied, there is no existing data set available to use directly for evaluation. We thus opted to create new data sets for evaluation. Table \ref{tab_sample_data} shows sample data from these datasets.

\textit{Standard Science Analogies (\textsc{std}):} As far as we could find, the closest dataset consisting of conceptual analogies is from \cite{turney2008latent}. It consists of ten standard science analogies. However, these only contain the source and target concepts but not any explanation in natural language.

\textit{\underline{S}cience \underline{a}nalogies from academic \underline{Q}\&\underline{A} sites \textsc{saqa}}: We searched for quiz questions that asked to create analogies on academic Q\&A sites like Chegg.com, Study.com \footnote{https://chegg.com/, https://study.com/. We manually inspected the data and found no personal identifiers or offensive content. We manually compiled the datasets, no scraping was done.} by using search queries like `create an analogy', `analogy to explain', and manually downloaded the relevant questions and answers. After manually removing irrelevant data, 75 unique question-answer pairs were obtained. Next, we manually extracted the analogies from answers, i.e., target and source concepts, and the explanation of the analogical similarity. 

There are total 109 concepts (about high-school science) with 148 English analogies. The average word length of analogies is 62.25 words.

% \jx{Table \ref{stats} can be removed and the numbers 
% can be described in words. For example, ``There are total 109 concepts with 148 analogies. The average word length of analogies
% is 62.25 words.''}.
%[Bhavya: Done, thanks.]

% \begin{table}[ht]
% \centering
%     \caption{\textsc{saqa} dataset statistics.}
%     \begin{tabular}{c|c|c | c}
%       \# Cncpts&  \# Analogies & \# Avg. words \\
%           \hline
% 109 & 148 & 65.25 \\
%     \end{tabular}
%     \label{stats}
% \end{table}

 \begin{table}[htbp]
 
    \centering
     \caption{Sample analogies from \textsc{std} and \textsc{saqa}.}
 \begin{tabular}{p{1cm}| p{0.8cm}| p{1.3cm} |p{2.8cm}}
Dataset &Target & Source & Explanation \\
 \hline
 \textsc{std} & atom & solar system & -\\
 \hline
\textsc{saqa} & ligase & sewing machine	& ... Ligase is similar to a sewing machine, as it binds two elements ... \textbf{(25 words)} \label{tab_sample_data}
% \vspace{-1mm}
  \end{tabular}
    \end{table}

\section{Experiment Results}

In this section, we present our experiment results and examine each of the three research questions introduced earlier. 

\subsection{Feasibility Analysis}
    \label{feasibility}
We first examine RQ1 and investigate whether InstructGPT is capable of generating analogies with simple prompts by 
%and identify the challenges associated with this task.  we first 
looking at the results on the smaller \textsc{std} dataset which contains well-known analogies. Here, we seek standard analogies, so we designed prompts with keywords such as "well-known analogy", "often used to explain", etc. The full list of prompts is in Table \ref{table: prompts_science}, Appendix \ref{appstd}).

We observed that all the prompts were successful in retrieving natural language analogies to some extent but they differed in several aspects. Table \ref{tab_sample_std} shows sample analogies generated by two of our prompts (P7 and P2, Table \ref{table: prompts_science}) for the target concept ``natural selection.'' In this case, the reference answer in the \textsc{std} dataset is ``artifical selection,'' which P7 successfully retrieved, while P2 generated a different but also valid analogy. Such variations indicate both the potential of using different prompts to generate (multiple) different analogies and the model sensitivity to prompt design, which we further investigate in Section \ref{sec_hypsen}. 

To quantify the effectiveness of different prompts, we manually evaluated the source concepts mentioned in the generated analogies (if any).  Table \ref{table: science} shows the number of exact matches of generated source concepts to those in the reference \textsc{std} dataset, along with the number of ``valid'' source concepts generated. Valid means a reasonable analogy that is either commonly known (e.g., easily available on the internet \footnote{Note that commonly known does not necessarily mean available on the internet. We use it only as a proxy here since there is no good way to determine what is common knowledge.}) or contains a meaningful justification. All prompts generated valid analogies in most cases, even if they didn't exactly match the reference source concept further suggesting the promise of InstructGPT for generating meaningful analogies. Note that the low number of exact matches with the reference dataset is expected to some extent because there are several possible ``valid'' analogies for a given source concept and so there is a small chance that the model would generate exactly the same analogous concept as in the reference.

% This suggests that a concept could have several valid analogies and it might be infeasible to pre-specify all the valid analogies, making it challenging to accurately evaluate such generated analogies without relying on manual assessment of each result.
% We will explore automatic evaluation in Sections \ref{auto_eval}, \ref{conscore}.

\begin{table}[h]
\centering
\caption{Number of analogies that match the ground truth or are otherwise meaningful, out of the total ten analogies generated for \textsc{std} target concepts by the seven prompts (P1-P7).}
\begin{tabular}{ c| c| c |c |c|c |c|c}
& P1 & P2 & P3 & P4 & P5 & P6 & P7\\
\hline
    \# Match & $3$ & $3$ & $6$ & $4$ & $3$ & $5$ & $3$\\
    \# Valid & $6$ & $9$ & $9$ & $8$ & $7$ & $10$ & $10$
    \label{table: science}
\end{tabular}
\vskip-15pt
\end{table}

%\subsection{Sensitivity to parameters}
%/subsection{Comparative analyses of prompts and temperature}
\subsection{Robustness analyses}
\label{sec_hypsen}
As observed in many other applications of prompted PLMs, the performance of a task tends to be sensitive to the prompts used and the temperature parameter \cite{lu2021fantastically, zhao2021calibrate}. Moreover, many PLMs are known to be vulnerable to the presence of spelling errors \cite{pruthi2019combating, ma2020charbert}. Thus, it is important to experiment with variations of both the prompts and the temperature parameter (with frequency\_penalty, Section \ref{temp} ), and spelling errors and 
study how they impact the generated analogy (RQ2). 
%their effectiveness. It is also known In general, if we view the prompts used as parameter values, we want to In our study, we experimented with two main types of parameters -- prompts and temperature (with frequency\_penalty, Section \ref{temp} ). In this section, we discuss how those impact the generated text. 

For these analyses, we need to compare the model performance in a large number of configurations, which makes human evaluation impossible. Thus, we rely on automatic metrics. Automatic evaluation of natural language generation is known to be challenging (e.g., long-form question answering \cite{krishna2021hurdles}) and automatic metrics 
%do not accurately reflect semantic similarity 
generally have low correlation with human judgment  \cite{callison2006re,raffel2019exploring}. Evaluation of analogies is even more challenging especially because a target concept could have several valid analogies with seemingly different meanings (e.g., ``artificial selection'' vs. ``sieve'' from Section \ref{feasibility}). Thus, before using existing methods, we designed sanity checks and found that those methods behave as we expect (e.g., analogies have a higher score than non-analogies, see Appendix \ref{auto_eval}). 
We note that our sanity checks are only the necessary and not the sufficient requirements of a good metric for evaluating analogies as they do not evaluate creativity or reasoning. However, we use them as an approximation only for relative comparison between methods on the same task as they are unlikely to favor any single method.
% For example, such automatic methods would be unable to effectively evaluate completely original and creative analogies, which is requires further research.

We use three representative measures of automatic evaluation of generated text: \textsc{bleurt} \cite{sellam2020bleurt} (\textsc{b}), \textsc{meteor} \cite{lavie2007meteor} (\textsc{m}), \textsc{rouge-l} (\textsc{r})\footnote{https://pypi.org/project/rouge-score/} \cite{lin2004rouge}\footnote{https://www.nltk.org/api/nltk.translate.meteor\_score.html}. \textit{\textsc{bleurt (b)}} is used as the primary metric for evaluation since it is a recent machine learning-based metric that has been shown to capture semantic similarities between texts \cite{sellam2020bleurt}.

Similar average \textsc{bleurt} values would indicate that the prompts are equally good (or bad) on a task, but not necessarily in the same way. On the other hand, Kendall's Tau \cite{kendall1938new} indicates how well the ranks of two variables are correlated. This would suggest that those prompts have similar strengths and weaknesses. Thus, we analyze both scores to get a more complete picture of hyperparameter sensitivity. 

% In our case, a high Kendall's Tau would indicate that the two prompts generally perform better (or worse), with possibly different magnitudes, on the same individual test samples. 

% \begin{figure}[h!]
%   \includegraphics[width=250pt]{figures/rougebasic.png}
%   \caption{Kendall's Tau correlation between \textsc{rouge-l} scores of various prompts and temperature values of the NO\_SRC settings}
%   \label{fig:rwosrc}
% \end{figure}

\subsubsection{Analysis of prompts}
\label{prompt_design}
%Previous work has shown that PLMs are sensitive to prompt design \cite{lu2021fantastically,zhao2021calibrate}. 
To study the effectiveness and robustness of different prompts for analogy generation in the unsupervised setting, we manually designed several prompts for all the problem settings. The different prompt variants are all paraphrases that are semantically similar. The main ways they differ are: 1. \textit{Questions vs. Imperative Statements} (e.g., P5 vs. P2, Table \ref{table: prompts_WSRC}); 2. \textit{Synonyms} (e.g., P2 vs. P3, Table \ref{table: prompts_WSRC});  3. \textit{Word Ordering} (e.g., P1 vs. P3, Table \ref{table: prompts_wosrc}). We only study the zero-shot setting mainly because the choice/number of examples in few-shot could make an impact on the generated analogies and make it harder to interpret our experiment results. 
 
\begin{table}[h]
    \caption{Prompts for \textsc{no\_src}}
    \begin{tabular}{p{.05\columnwidth}|p{.82\columnwidth}}
        \textbf{Id} & \textbf{Prompt} \\
           \hline
       P1 & Explain <target> using an analogy. \\
\hline
P2 & Create an analogy to explain <target>.\\ 
\hline
P3 & Using an analogy, explain <target>.\\
\hline
P4 & What analogy is used to explain <target>?\\
\hline 
P5 & Use an analogy to explain <target>.
    \end{tabular}
    \vspace{-2mm}
    \label{table: prompts_wosrc}
\end{table}

% Although recent work has shown promise in automatically learning prompts, there is no existing dataset of analogies in natural language, so we only study prompts in the unsupervised setting. 
Prompts for the \textsc{no\_src} and \textsc{wsrc} settings are in Tables \ref{table: prompts_wosrc},\ref{table: prompts_WSRC}, respectively. Here, <target>, <src> are target and source concept placeholders.

\begin{table}[h]
    \caption{Prompts for \textsc{wsrc}}
    \begin{tabular}{p{.05\columnwidth}|p{.82\columnwidth}}
        \textbf{Id} & \textbf{Prompt} \\
           \hline
       P1 & Explain <target> using an analogy involving <src>. \\
\hline
P2 & Explain how <target> is analogous to <src>.\\ 
\hline
P3 & Explain how <target> is like <src>.\\
\hline
P4 & Explain how <target> is similar to <src>.\\
\hline
P5 & How is <target> analogous to <src>?\\
\hline
P6 & How is <target> like <src>?\\
\hline
P7 & How is <target> similar to <src>?
    \end{tabular}
\vspace{-3mm}
    \label{table: prompts_WSRC}
\end{table}

Our major findings are as follows:

\textit{Questions and statements are significantly different}: The question prompts are P4, Table \ref{table: prompts_wosrc} and P5-P7, Table \ref{table: prompts_WSRC}. From Tables  \ref{tab_wosrc} and \ref{tab_WSRC}, questions have significantly different and lower scores than statements. This could be an artifact of how the InstructGPT models were trained and should be further investigated.

\begin{table}[h]
\centering
\caption{Comparison of performances of different prompts and temperatures in \textsc{no\_src}. $*$ and $\dagger$ mean statistically significant compared to the best performing setting at p<0.1 and p<0.05 respectively based on a two-tailed t-test.}
\begin{tabular}{ c| c| c |c }
& B & R & M \\
\hline
P1$_{tl}$ &  $0.46$ & $0.187$ & $0.154$ \\
P1$_{th}$ & $0.448^{\dagger}$  & $0.181^{\dagger}$ & $0.167$ \\
P2$_{tl}$ & $0.451$ & $0.193$ & $0.154$ \\
P2$_{th}$ &  $0.45^{*}$ & $0.184$ & $0.161$ \\
P3$_{tl}$ & $\textbf{0.462}$ & $\textbf{0.196}$  & $0.164$ \\
P3$_{th}$ & $0.452$ & $0.188$  &  $\textbf{0.171}$\\
P4$_{tl}$ & $0.427^{\dagger}$ & $0.170^{\dagger}$  & $0.126^{\dagger}$ \\
P4$_{th}$ & $0.431^{\dagger}$ & $0.179^{\dagger}$  & $0.156$ \\
P5$_{tl}$ & $0.451$ & $0.188$ & $0.154$ \\
P5$_{th}$ & $0.449^{*}$ & $0.183^{*}$ & $0.163$ 
\vspace{-1em}
 \label{tab_wosrc}
\end{tabular}
\end{table}

% as follows: P4 in \textsc{no\_src} and P5-P7 in \textsc{wsrc}. From Figure \ref{fig:bwosrc} top-left and bottom-right, as expected, those questions and statements that share the most words (e.g., P3 and P6) are more strongly correlated. Even so, f

\textit{Impact of synonyms and word order:} Prompt performances vary based on synonyms and word order. For example, some synonymous prompt pairs (e.g, P2-P4, P5-P7 in \textsc{wsrc}) are more correlated than others (e.g., P2-P3, P5-P6 in \textsc{wsrc}). This could be because ``analogous to'' and ``similar to'' share a word unlike the other synonym ``like''. As expected, prompts with the most different meanings (e.g., P1 in \textsc{wsrc} -- involving <src> is not necessarily the same as analogous to <src>) are least correlated with others. However, from Table \ref{tab_WSRC}, the average performances of synonymous prompts (e.g., $P2_{tl}$ and $P3_{tl}$, $P2_{tl}$ and $P5_{tl}$) are not significantly different. Overall, this suggests that InstructGPT is more robust to synonyms/word-order than to the prompt style (question/imperative statements) for this task. 
% \jx{Did you run statistical test to draw this conclusion? If so, please add some discussion on it}.
The overall best-performing prompts (P3 in \textsc{no\_src}, P2 in \textsc{wsrc}) contain some form of the word ``analogy'' rather than its synonyms, confirming that precise and direct prompts are better.

\begin{table}[htbp]
\centering
  \caption{Comparison of performances of different prompts and temperatures in \textsc{wsrc}. $*$ and $\dagger$ mean statistically significant at p<0.1 and p<0.05 compared to the best performing setting respectively based on a two-tailed t-test.}
\begin{tabular}{ c| c| c |c }
& B & R & M \\
\hline
P1$_{tl}$ &  $0.504$ & $0.223$  & $0.187^{\dagger}$ \\
P1$_{th}$ & $0.497^{\dagger}$ & $0.212^{\dagger}$ & $0.199$  \\
P2$_{tl}$ & $\textbf{0.515}$ & $0.217$  & $0.203$ \\
P2$_{th}$ & $0.502^{*}$ & $0.210^{\dagger}$ & $\textbf{0.208}$  \\
P3$_{tl}$ & $0.504$ & $\textbf{0.229}$  & $0.191$ \\
P3$_{th}$ & $0.504$ & $0.216$  & $0.203$  \\
P4$_{tl}$ & $0.506$ & $0.214$ & $0.197$ \\
P4$_{th}$ & $0.497^{\dagger}$ & $0.206^{\dagger}$ &  $0.2$ \\
P5$_{tl}$ &  $0.499^{*}$ &  $0.217$ & $0.18^{\dagger}$ \\
P5$_{th}$ & $0.496^{\dagger}$ & $0.211^{\dagger}$ &  $0.191^{*}$ \\
P6$_{tl}$ & $0.500^{*}$ & $0.216$  & $0.176^{\dagger}$ \\
P6$_{th}$ & $0.494^{\dagger}$ & $0.212^{\dagger}$ & $0.183^{\dagger}$ \\
P7$_{tl}$ & $0.497^{\dagger}$ & $0.208^{\dagger}$  & $0.179^{\dagger}$ \\
P7$_{th}$ & $0.492^{\dagger}$ & $0.204^{\dagger}$ & $0.186^{\dagger}$
\vspace{-2mm}
 \label{tab_WSRC}
\end{tabular}
\end{table}

% We further inspect an instance where \textsc{bleurt} correlation scores are low ($P3_{tl}$  -- the overall best performing prompt and $P4_{tl}$). One of the least correlated scores for these prompts are 0.479 and 0.629 respectively. In this case, $P4_{tl}$ generated \textit{``A chloroplast is like a small factory inside a plant cell. It makes food for the plant using sunlight and water.''} and $P3_{tl}$ generated \textit{``Chloroplasts are like the lungs of a plant. They take in carbon dioxide from the air and release oxygen. They also use sunlight to convert carbon dioxide and water into glucose and oxygen.''} Both the analogies are valid, however, the first matches the human reference in our dataset, \textit{``Chloroplast is like the kitchen of the factory where food is synthesised.''} 

% \begin{figure}[h!]
%   \includegraphics[width=250pt]{figures/rougesrcgiven.png}
%   \caption{Kendall's Tau correlation between \textsc{rouge-l} scores of various prompts and temperature values of the \textsc{wsrc} settings}
%   \label{fig:rWSRC}
% \end{figure}
\setlength{\belowcaptionskip}{-10pt}

\begin{figure}[h!]
  \includegraphics[width=200pt]{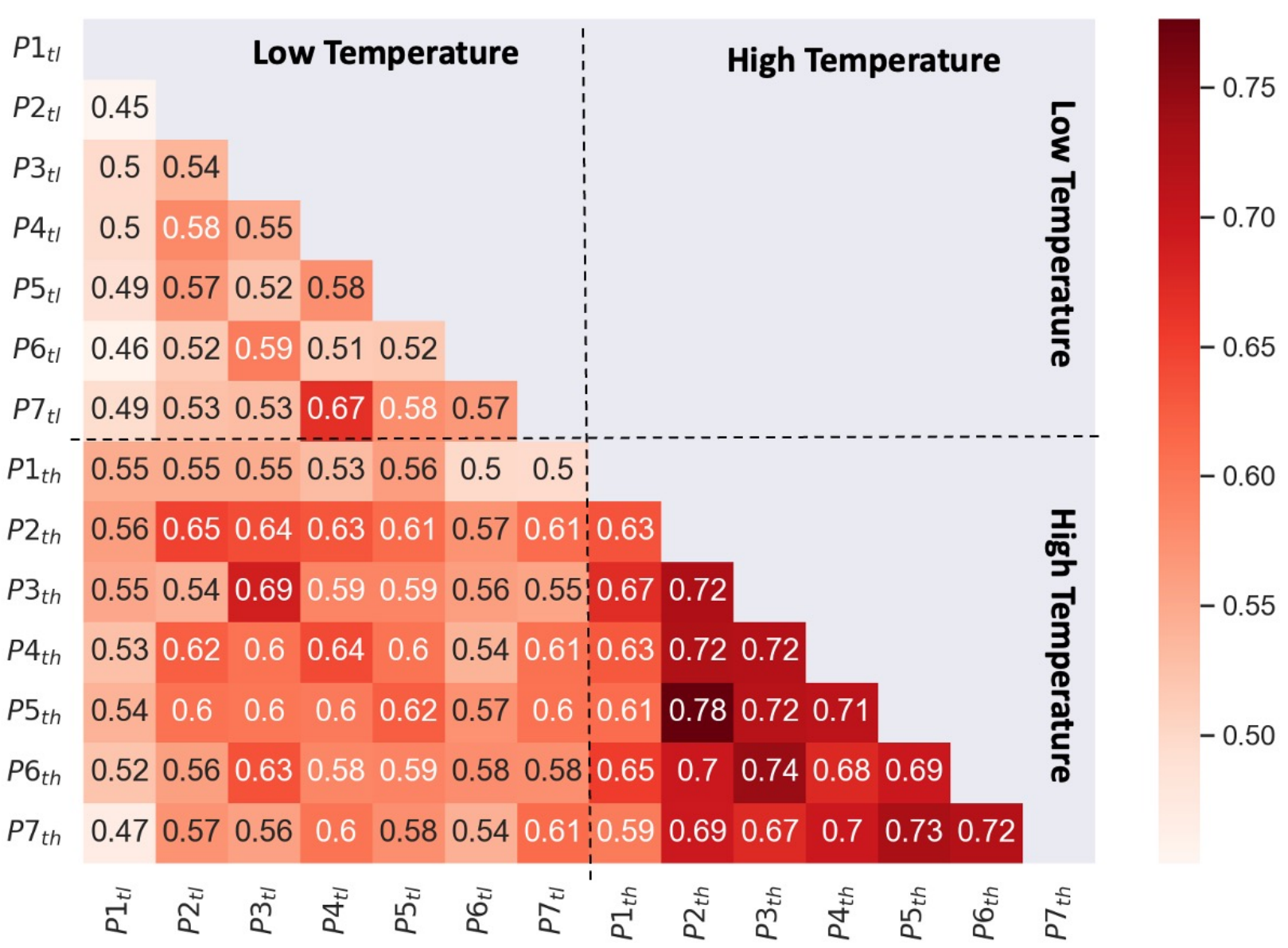}
  \caption{Kendall's Tau correlation between \textsc{bleurt} scores of various prompts and temperatures in \textsc{wsrc}}
  \label{fig:bWSRC}
\end{figure}

\subsubsection{Analysis of temperature}
\label{temp}
Higher temperature increases the randomness in the generated text and is often suggested for creative tasks \cite{lucy2021gender}. Since some analogies require creativity, we are especially interested in studying the impact of this hyperparameter.

We explore two settings. \textbf{Low Temperature (tl)}: this is a deterministic setting, where temperature = frequency\_penalty = presence\_penalty = 0. \textbf{High Temperature (th)}: Here temperature is set to 0.85. To avoid repetition of words and topics, we set frequency\_penalty = 1.24 and presence\_penalty = 1.71. These hyperparameters were selected based on initial qualitative exploration. To account for the randomness, we set best\_n = 3, i.e., select the best response out of three generated responses, and generate 5 such best responses. In all experiments, we report the average performance of all 5 responses. 

From the results in lower-right corner of Figures \ref{fig:bWSRC} and \ref{fig:bwosrc} (in Appendix \ref{saqaappendix}), we make the following two observations:

Firstly, at high temperature, prompts are generally well-correlated with each other suggesting lesser sensitivity to prompt design at high temperatures. This requires further investigation because we expect higher randomness to generate a variety of different analogies, and thus have lower correlations in general.
% \cz{why is this expected? High randomness would lead to a variety of different analogies being generated, but it's unclear why the correlation should be higher in such a case as compared with a low temperature, where we should expect to see a more consistent analogy being generated.}

Secondly, the overall performances of the high-temperature variants are generally lower than their low-temperature counterparts. To investigate when high temperature could help, we further looked into a case in the \textsc{wsrc} setting where the high-temperature version of the best prompt, ($P2_{th}$), performed much better. The results are shown in Table \ref{tab_temp_example}. In this case, unlike $P2_{hl}$, $P2_{tl}$ fails on identifying the target and also generates incorrect facts, (``rubber of your lungs''). This shows some evidence of high temperature prompts working better for more complex and creative analogies, which should be investigated further in the future.

\begin{table*}[h]
    \centering
     \caption{Analogies generated in the high and low-temperature setting by the prompt: \textit{Explain how breathing mechanism of frogs is analogous to blowing of the balloon.}
  \label{tab_temp_example}}
\begin{tabularx}{\linewidth}{|c|L|} 
\hline
 \textbf{Temp.} & \textbf{Generated Analogy}  \\
 \hline
\textbf{High}& ... When you blow into a balloon, the air pressure ... The frog's respiratory system works in a similar way. When you breathe in, you increase the air pressure inside your lungs. This increased air pressure causes the walls ... \\
    \hline
\textbf{Low}  & When you blow up a balloon, the air inside the balloon is pushing outwards on the rubber of the balloon. This is the same thing that happens when you breathe in: the air inside your lungs is pushing outwards on the rubber of your lungs.\\
\hline
\end{tabularx}
    \end{table*}

\subsubsection{Analysis of synthetic spelling errors}
Since obtaining natural spelling errors can be difficult, injecting synthetic errors is a common way to test the robustness of models (e.g., \cite{jayanthi2020neuspell}. Thus, following previous work \cite{sakaguchi2017robsut, jayanthi2020neuspell}, we injected the following four types of character-level errors to the internal characters of the target concept in the prompt: Delete (delete one randomly chosen character), Permute (switch two randomly chosen adjacent characters in the string), Insert (insert one random alphabet at a random position), and Replace (replace one randomly chosen character in the string with a random alphabet). Target concepts with length less than 3 were kept unchanged. 

Average \textsc{bleurt} scores from three different runs for all prompts in the low-temperature setting in \textsc{no\_src} are shown in Table \ref{table: spell}. Overall, the performance decreases, indicating the sensitivity of language models to spelling errors. Further, Replace generally leads to the biggest performance drop for all prompts ($\sim 3-7\%$ relative decrease). The model is generally most robust to Insert, similar to the results reported in previous work on word recognition using neural networks \cite{sakaguchi2017robsut}.

% Further, based on observation, we found that performance is generally worse for noise added to shorter strings (e.g, d\textbf{n}a $\rightarrow$ d\textbf{l}a) compared to longer ones (e.g., tu\textbf{m}or suppressor genes $\rightarrow$ tu\textbf{p}or suppressor genes). This is expected because the model could struggle more when there is a higher percentage of incorrect characters in the string.

\setlength{\tabcolsep}{5pt}
\begin{table}[h]
\centering
\caption{Impact of injecting Delete (D), Permute (P), Insert (I) and Replace (R) errors to the target concept in the prompt compared to the original (O) prompt based on \textsc{bleurt} scores. $*$ and $\dagger$ mean statistically significant at p<0.1 and p<0.05 respectively based on a two-tailed t-test.}
\begin{tabular}{ c| c| c |c |c|c }
 & D & P & I & R & O\\
\hline
P1 & $0.438^{\dagger}$ & $0.437^{\dagger}$ & $0.436^{\dagger}$ & $0.429^{\dagger}$ & $0.46$ \\
P2 & $0.431^{\dagger}$ & $0.434^{\dagger}$ & $0.442$ & $0.427^{\dagger}$ & $0.451$\\
P3 & $0.444^{\dagger}$ & $0.445^{\dagger}$ & $0.447^{*}$ & $0.44^{\dagger}$ & $0.462$\\
P4 & $0.423$ & $0.424$ & $0.428$ & $0.416$ & $0.427$\\
P5 & $0.438^{*}$ & $0.437^{*}$ & $0.441$ & $0.435^{\dagger}$ & $0.451$
    \label{table: spell}
    \vspace{1mm}
\end{tabular}
\end{table}

\subsection{Analysis of model size}

Finally, we examine RQ3, i.e., how does the model size impact the quality of the generated analogies. In general, models with more parameters can be expected to perform better. We now study whether the same holds for this task and how much the model size impacts the performance. 

\begin{figure}[h!]
  \includegraphics[width=215pt]{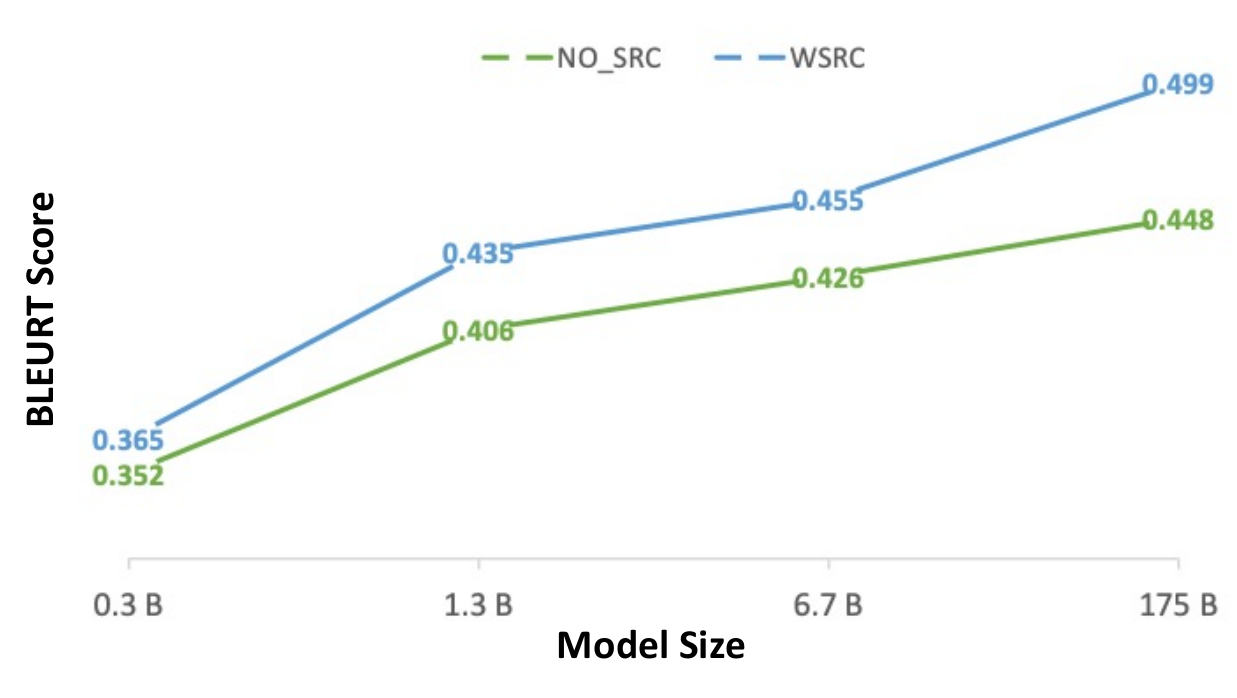}
  \caption{Average performances of various InstructGPT models based on \textsc{bleurt} scores.}
  \label{fig:mod_cmp}
\end{figure}

Figure \ref{fig:mod_cmp} shows the \textsc{bleurt} scores of various models on both the task setups. As expected, the performance increases significantly with model size in both \textsc{wsrc} and \textsc{no\_src}, suggesting that larger models are better at generating analogy-like text for the given targets. Further, the biggest improvement is seen as the number of parameters increases from 0.3B to 1.3B in both settings (19.17\% and 15.34\% relative improvements, respectively).
% not for the biggest increase in number of paprameters (6.7 B to 175 B, 9.67\% and 5.16\% relative improvements, respectively). 

Similar to what we observed in the case of the 175B Davinci model, the performance in \textsc{wsrc} is higher than that in \textsc{no\_src} for other models too. This confirms that all models have some capacity to incorporate the source provided in the prompt.

\subsection{Human evaluation}

To further validate the generated analogies more comprehensively, we also conducted human evaluation as described below.
 
\subsubsection{Annotation Setup}
We conducted the study on Amazon Mechanical Turk. Based on manual evaluation of responses to screening tests (Appendix \ref{mturk}), we selected 17 workers for the main study.

Further, we created a sample dataset for evaluating analogies generated both in the \textsc{no\_src} and  \textsc{wsrc} settings. In total, we generated  13k analogies \footnote{6 analogies (5 in high temperature and 1 low temperature) *109 target concepts*5 prompts*4 models} in \textsc{no\_src} and 18k analogies \footnote{6 analogies (5 in high temperature and 1 low temperature) *109 target concepts*7 prompts*4 models} in \textsc{wsrc}. From this data, we randomly selected 42 concepts for the \textsc{no\_src} setup and 21 of them were selected for the \text{wsrc} setup (to have comparabale number of analogies in both settings). The analogies for the selected concepts, generated by all the models using all the prompts in the low temperature setting were selected for evaluation since low temperature was better based on automatic evaluation. 
% Further, we sampled less concepts in the \textsc{wsrc} setting concepts to have similar number of analogies annotated for each setting since \textsc{wsrc} had more prompts. 

In total, 1407 unique analogies (576 from \textsc{wsrc}, 770 from \textsc{no\_src}, and 61 human-generated from \textsc{saqa}) were evaluated by 3 workers each, which is common in previous work on evaluation of automatically generated text \cite{van2021human}. The main study had one question asking workers to evaluate whether the shown candidate analogy was meaningful for the target concept (Yes/No/Can't decide) and provide a text input for explaining their choice (Figure \ref{fig:int_main}, Appendix \ref{interface}). Please refer to Appendix \ref{mturk} for more details of the study design.

\subsubsection{Quantitative Results}
Table \ref{hum_eval} shows the percentage of analogies rated as meaningful, based on majority vote, for the various models and the human references from \textsc{saqa}. There were <2\% ties or cases with `Can't decide' as the majority, which were discarded. The Fleiss' kappa \cite{fleiss1971measuring} inter-annotator agreement was 0.347 in case of \textsc{wsrc} (plus human references for the selected concepts for \text{wsrc} concepts), indicating fair agreement and 0.553 in case of \textsc{no\_src} (plus human references for the selected concepts for \textsc{no\_src} concepts) indicating moderate agreement. 

We observe that the percentage of meaningful analogies increases with model size, again confirming that larger models have a higher capacity to generate analogies. Interestingly, in the \textsc{no\_src} setting, the largest model has comparable performance to humans. We note that this doesn't necessarily mean that those models are creative or have commonsense reasoning skills as they could have simply memorized those analogies, which a known problem of such models \cite{bender2021dangers}. It requires further research to test whether the models generate novel analogies unseen during training.  

Moreover, upon inspection, we found that the human-generated analogies sometimes had minor issues, such as grammatical errors, which could impact their rating by annotators. So, it is possible that analogies written by experts, such as science instructors proficient in English, might be rated higher. Nevertheless, these results are quite encouraging as the model seems to have comparable performance to general online users who wrote the analogies in our reference dataset.

In the \textsc{wsrc} setting, the performance of InstructGPT is lower than human performance. This could be because there is a lesser likelihood of seeing the exact same analogy, i.e., the one asked to explain in the prompt, during training, compared to seeing \textit{any} analogies for the target concept as required in the \textsc{no\_src} setting. So, \textsc{wsrc} might require more ``analogical reasoning'' from the models, especially for explaining analogies not seen during training. This highlights the importance of human evaluations for such tasks because otherwise, based on automatic evaluation alone, we would conclude that this is an easier setting. This is because metrics like \textsc{bleurt} cannot assess the soundness of the generated reasoning.

We also compute the \textsc{no\_src} performance on  the 21 shared concepts (\textsc{no\_src\textsubscript{21}}, Table \ref{hum_eval}) for a fair comparison between the two settings. It is interesting to note that the performances of smaller models increase while that of larger models go down in the \textsc{wsrc} setting. This could be because the provided source in the prompt helps provide some guidance to the smaller models. For example, even by copying parts of the prompt (i.e., source and target), they could generate meaningful analogies (e.g., <source> is like <target>) in a few cases. Since their performance in the \textsc{no\_src} setting is very poor, even minor help or ``tricks'' would lead to performance improvement. On the other hand, the larger models that already performed very well, likely do not have much to gain from such help and, in fact, perform worse due to the analogical reasoning argument made above. 

Overall, this highlights some limitations of the InstructGPT model for analogical reasoning, which requires further research for improvement.

\subsubsection{Error Analysis}

The annotators were also asked to explain their answer choice (i.e, meaningful analogy or not). By inspection, we identified the following major themes based on the workers' explanations for choosing ``not meaningful'' across all models/tasks. These themes are not mutually exclusive and multiple themes were often found for one wrong generation. 

\textbf{1. No Analogy:} This is one of the most common cases where the model failed to generate any analogy at all. Instead, it mostly generated a simple description/definition of the target concept. In a few cases, it also generated a tautology or an example. For example, \textit{``The b-lymphocytes are similar to the white blood cells.''}

\textbf{2. Irrelevant to target:} The generated text contained little to none relevant information pertaining to the target. One interesting reason behind this was capitalization for abbreviations. For example, since the targets in the prompt were lowercased (e.g., nadh), smaller models were unable to identify abbreviations, while the larger models succeeded at this. Another reason observed was that of an ambiguous target, e.g., computer ``mouse'' misidentified as a rodent. In more insidious cases, the text looked correct but presented incorrect facts. 

\textbf{3. Incorrect source or explanation:} 
Here, important details about the source concept were either incorrect or missing, or the provided explanation was insufficient, making the analogy completely wrong or weak at best. For example, \textit{``A molecule of DNA is like a drop of water. It has a specific shape and size, and it can carry the genetic instructions for making a particular organism.''}

Some error types found in other natural language generations from GPT-3 \cite{dou2021scarecrow}, e.g., incoherence and grammar, were also found in our task. Further research is required to quantify them for analogical generation and attempt to fix them. 

\begin{table}[t]
\centering
  \caption{Percentage of meaningful analogies generated by various InstructGPT models and humans based on human evaluation. Highest value per row is underlined.}
  \setlength\tabcolsep{4pt} % default value: 6pt
  \begin{tabular}{c|c|c|c|c|c}
& 0.3B &  1.3B & 6.7B & 175B & Human \\ 
\hline
 \textsc{no\_src} & 1.90 & 15.61 & 48.29 & \underline{70.05} & 66.67 \\
 \textsc{wsrc}  & 8.97 &  29.05 & 38.46  & 53.79 & \underline{71.88}\\ 
 \textsc{no\_src\textsubscript{21}} & 0 & 12.0 & 47.0 & 66.99 & \underline{71.88} 
 \label{hum_eval}
 \vspace{-6mm}
\end{tabular}
\end{table}

%% file: appendix.tex
\appendix

\section{Hyperparameters}
\label{hyperparam}
Based on initial explorations, where we varied the number of maximum tokens between 0 and 1000 in increments of 100, and then from 935-955 in increments of 1, we noticed that setting a high number of maximum tokens worked better in generating more comprehensive analogies that were not abruptly cut-off and there was little sensitivity to higher values around 950.
So, we randomly chose one value in that range (939). The default value of top\_p = 1 was used.

\section{Suitability of existing evaluation metrics}
\label{auto_eval}

To first investigate the suitability of existing evaluation metrics for generated analogies before we can trust any evaluation results using them, we designed two testers to examine whether the existing metrics behave as expected: 1) {\bf Ordering Tester \textsc{ot}}: This tester is to see if an evaluation metric 
can order a set of methods that have known orders between them correctly as expected. 
2) {\bf Random Perturbation Tester \textsc{rpt}} : This tester checks if an evaluation metric responds to a random perturbation to the ground truth data used for evaluation. A reasonable metric is expected to generate lower performance figures after perturbation. 

%to have a particular order among them in terms of performance. 
%We aim to study whether existing metrics are capable of evaluating analogies as expected. In other words, do they 
%order the methods, which are expected to perform worse, lower than others? 

%Secondly, we investigate the discerning power of the metrics, i.e., how capable are they in distinguishing between baselines and various settings. Intuitively, a good metric should give much lower scores to random baselines.

We use those two testers to study the suitability of three popular and representative measures of automatic evaluation of generated text: \textsc{bleurt} \cite{sellam2020bleurt}, \textsc{meteor} \cite{lavie2007meteor}, \textsc{rouge-l} \cite{lin2004rouge}. 

\textit{\textsc{bleurt (b)}} is a recent machine learning-based metric that has been shown to capture semantic similarities between text. \textit{\textsc{rouge-l (r)}}\footnote{https://pypi.org/project/rouge-score/} measures longest matching subsequence of words. We use its F1-score. \textit{\textsc{meteor (m)} }\footnote{https://www.nltk.org/api/nltk.translate.meteor\_score.html} matches word stems and synonyms also.

%\subsubsection{Order and Discerning Power Checks}
%\subsubsection{Design of testers}
\noindent {\bf Design of testers:}
We design an \textsc{ot} and a \textsc{rpt} based on the following baseline methods: 
%To define an expected ordering of methods, we set up two baselines described below.

\textit{No Analogy baseline (\textsc{no\_anlgy}):} Here, the prompts instruct the model to generate an explanation or description of the target concept and do not ask for an analogy explicitly. Thus, we expect the generated text to be in a different ``style'' than analogies and the overall performance to be lower. However, the generation would still contain other relevant keywords describing the target. Thus, it is a good baseline to test if the metrics can distinguish between analogies and other descriptions. 

\textit{Random baselines:} For each of the three setups, we introduced random baselines (\textsc{no\_anlgy\_rand}, \textsc{\textsc{no\_src}\_rand}, and \textsc{wsrc\_rand}, respectively) where a generated string is evaluated against a random analogy (excluding the correct matching analogy) in the reference dataset (i.e., applying a random perturbation to the ground truth). These baselines preserve the ``style'' of the text but not the content.  We expect these methods to perform worse than their non-random counterparts.

Additionally, \textsc{no\_src} setting is expected to perform worse than \textsc{wsrc} because in \textsc{wsrc}, the model has more information (i.e., the source concept) and thus has better chances of generating the correct analogical explanation. Thus, the expected order is \textsc{no\_anlgy} < \textsc{no\_src} < \textsc{wsrc}. 

%\subsubsection{Metric testing results}

\noindent {\bf Metric testing results:} Table \ref{tabmetrics} shows the overall results of experiments on the \textsc{saqa} dataset using the Davinci model. Each row shows the highest average scores given by a metric in various setups (performances of each prompt are in Section \ref{sec_hypsen} and at the end of this section.). 

We can see that all the three metrics order the setups as expected, i.e., random baselines are assigned a lower score than non-random setups, and scores for \textsc{no\_anlgy} < \textsc{no\_src} < \textsc{wsrc}. This suggests that all the three metrics have ``passed" our two testers and thus can be reasonably used to evaluate whether the automatically generated analogies are similar to those generated by humans. In other words, they should help assess whether the generated text is relevant to the target concept and discuss properties of the concept that could be explained using analogies (because they passed \textsc{rpt}), and written in an analogical style (because they passed \textsc{ot}). 

Moreover, the results also indicates that the InstructGPT model is able to follow the prompts in the three settings to some extent and generate non-analogical descriptions, general analogies, and analogies containing the source concepts, in those settings respectively. 

In terms of discernment power, all metrics have small gaps between the scores of random and non-random settings. Similar results were previously reported in \cite{krishna2021hurdles} for \textsc{rouge} scores on long-form question-answering. Out of the three metrics, the \textsc{bleurt} score has the largest gaps in all the settings, both between the random and non-random baselines and also between settings. It is also shown to capture semantic similarity well \cite{sellam2020bleurt}. Thus, we use it as the main metric in the rest of the experiments.   
% Recentl explanation \cite{clinciu2021study}

\begin{table*}
\centering
  \caption{Testing results using \textsc{ot} and \textsc{rpt}. The higher score between the random baseline and the non-random setup is bolded. Highest score in a row in underlined.}
  \begin{tabular}{c|c|c|c|c|c|c}
& \textsc{no\_anlgy\_rand} & \textsc{no\_anlgy} & \textsc{no\_src\_rand} & \textsc{no\_src} & \textsc{wsrc\_rand} & \textsc{wsrc}  \\ 
\hline
B & 0.349 & \textbf{0.445} & 0.375 & \textbf{0.462} & 0.385 & \underline{\textbf{0.515}} \\
R & 0.122 & \textbf{0.183} & 0.132 & \textbf{0.196} & 0.122 & \underline{\textbf{0.229}}  \\
M & 0.099 & \textbf{0.158} & 0.109 & \textbf{0.171} & 0.109 & \underline{\textbf{0.208}}
\end{tabular}
\label{tabmetrics}
\end{table*}

\begin{table}[H]
\centering
    \caption{Prompts for \textsc{no\_anlgy}}
    \begin{tabular}{p{.05\columnwidth}|p{.82\columnwidth}}
        \textbf{Id} & \textbf{Prompt} \\
           \hline
       P1 & Explain <target>. \\
\hline
P2 & What is <target>?\\ 
\hline
P3 & Explain <target> in plain language to a second grader.\\
    \end{tabular}
    \label{table: prompts_woanlgy}
\end{table}

\begin{table}[H]
\centering
\caption{Comparison of performances of different prompts and temperatures in \textsc{no\_anlgy}.}
\begin{tabular}{ c| c| c |c }
& B & R & M \\
\hline
P1$_{tl}$ & 0.434 & \textbf{0.183} & 0.149  \\
P1$_{th}$ & 0.432 & 0.18 & \textbf{0.158} \\
P2$_{tl}$ & 0.43 & 0.175 & 0.129\\
P2$_{th}$ & 0.425 & 0.172 & 0.136 \\
P3$_{tl}$ & \textbf{0.445} & 0.180 & 0.132  \\
P3$_{th}$ & 0.444 & 0.179 & 0.144
 \label{tab_woanlgy}
\end{tabular}
\end{table}

\begin{table}[H]
\centering
\caption{Comparison of performances of different prompts and temperatures in \textsc{no\_src\_rand}.}
\begin{tabular}{ c| c| c |c }
& B & R & M \\
\hline
P1$_{tl}$ & \textbf{0.375}  & \textbf{0.132} & 0.103 \\
P1$_{th}$ &  0.367  & 0.123  & 0.108\\
P2$_{tl}$ & 0.359 & 0.116 & 0.092 \\
P2$_{th}$ &  0.366 & 0.127 & 0.105 \\
P3$_{tl}$ & 0.362 & 0.124   & 0.099 \\
P3$_{th}$ & 0.364 & 0.126 & \textbf{0.109}   \\
P4$_{tl}$ & 0.338 & 0.115  & 0.084 \\
P4$_{th}$ & 0.348 &  0.121 &  0.1 \\
P5$_{tl}$ & 0.358 & 0.121 & 0.097 \\
P5$_{th}$ & 0.348 & 0.122 & 0.107
 \label{tab_wosrc_RAND}
\end{tabular}
\end{table}

\begin{table}[H]
\centering
\caption{Comparison of performances of different prompts and temperatures in wsrc\_rand.}
\begin{tabular}{ c| c| c |c }
& B & R & M \\
\hline
P1$_{tl}$ &  0.37 & 0.120  & 0.094 \\
P1$_{th}$ & 0.363 & \textbf{0.122} & 0.107 \\
P2$_{tl}$ & \textbf{0.385} &  0.117 & 0.096 \\
P2$_{th}$ & 0.381 & 0.12 & \textbf{0.109} \\
P3$_{tl}$ & 0.358 & 0.117 & 0.095 \\
P3$_{th}$ & 0.359 & 0.115 & 0.1 \\
P4$_{tl}$ &  0.367 & 0.113 & 0.096  \\
P4$_{th}$ & 0.37 & 0.115 & 0.105 \\
P5$_{tl}$ &0.36  & 0.113 & 0.09  \\
P5$_{th}$ & 0.356 & 0.117 & 0.094 \\
P6$_{tl}$ & 0.346  & 0.111 &  0.086 \\
P6$_{th}$ & 0.347 &  0.113 & 0.091 \\
P7$_{tl}$ &  0.353 &  0.114 & 0.092 \\
P7$_{th}$ & 0.352 & 0.109 & 0.093
 \label{tabNOSRCRAND}
\end{tabular}
\end{table}

\begin{table}[H]
\centering
\caption{Comparison of performances of different prompts and temperatures in \textsc{no\_anlgy\_rand}.}
\begin{tabular}{ c| c| c |c }
& B & R & M \\
\hline
P1$_{tl}$ & 0.346 & 0.115  & 0.087  \\
P1$_{th}$ & \textbf{0.349} & \textbf{0.122} & \textbf{0.099}  \\
P2$_{tl}$ & 0.322 & 0.116 & 0.077 \\
P2$_{th}$ & 0.327 & 0.113 &  0.081 \\
P3$_{tl}$ & 0.334  & 0.111 & 0.079 \\
P3$_{th}$ &  0.336 & 0.11  & 0.081
 \label{tab_woanlgy_RAND}
\end{tabular}
\end{table}

\section{Experiments on \textsc{std} dataset}
\label{appstd}
\begin{table}[H]
    \caption{Prompts for \textsc{std} analogies}
    \begin{tabular}{p{.05\columnwidth}|p{.82\columnwidth}}
        \textbf{Id} & \textbf{Prompt} \\
           \hline
       P1 & Explain <target> using an analogy. \\
\hline
P2 & Explain <target> using a well-known analogy.\\ 
\hline
P3 & What analogy is often used to explain <target>?\\
\hline
P4 & Using a well-known analogy, explain <target>.\\
\hline
P5 & Using an analogy, explain <target>.\\
\hline
P6 & What is a well-known analogy to explain <target>?\\
\hline
P7 & What is analogous to <target>?
    \end{tabular}
    \label{table: prompts_science}
\end{table}

\begin{table}[H]
\centering
\caption{Most common analogies generated for each target concept in the \textsc{std} dataset. \#Pmt. means number of prompts that generated the shown analogy.}
\label{tab_st_agt}
\begin{tabular}{ c| c| c}
Target & Most common src.& \# Pmt. \\
\hline
mind & computer&	7  \\
atom &	solar system & 6    \\
heat transfer &fluid/water flow &	4   \\
sounds & wave  & 4  \\
respiration & combustion & 3   \\
light & river &	3 \\
planet & rock	& 2  \\
bacterial mutation	& game of telephone & 3 \\
natural selection & sieve &	2 \\
gas molecules & balls &	2 \\
\end{tabular}
\end{table}

\section{Experiments on \textsc{saqa} dataset}
\label{saqaappendix}

\begin{figure}[h!]
  \includegraphics[width=200pt]{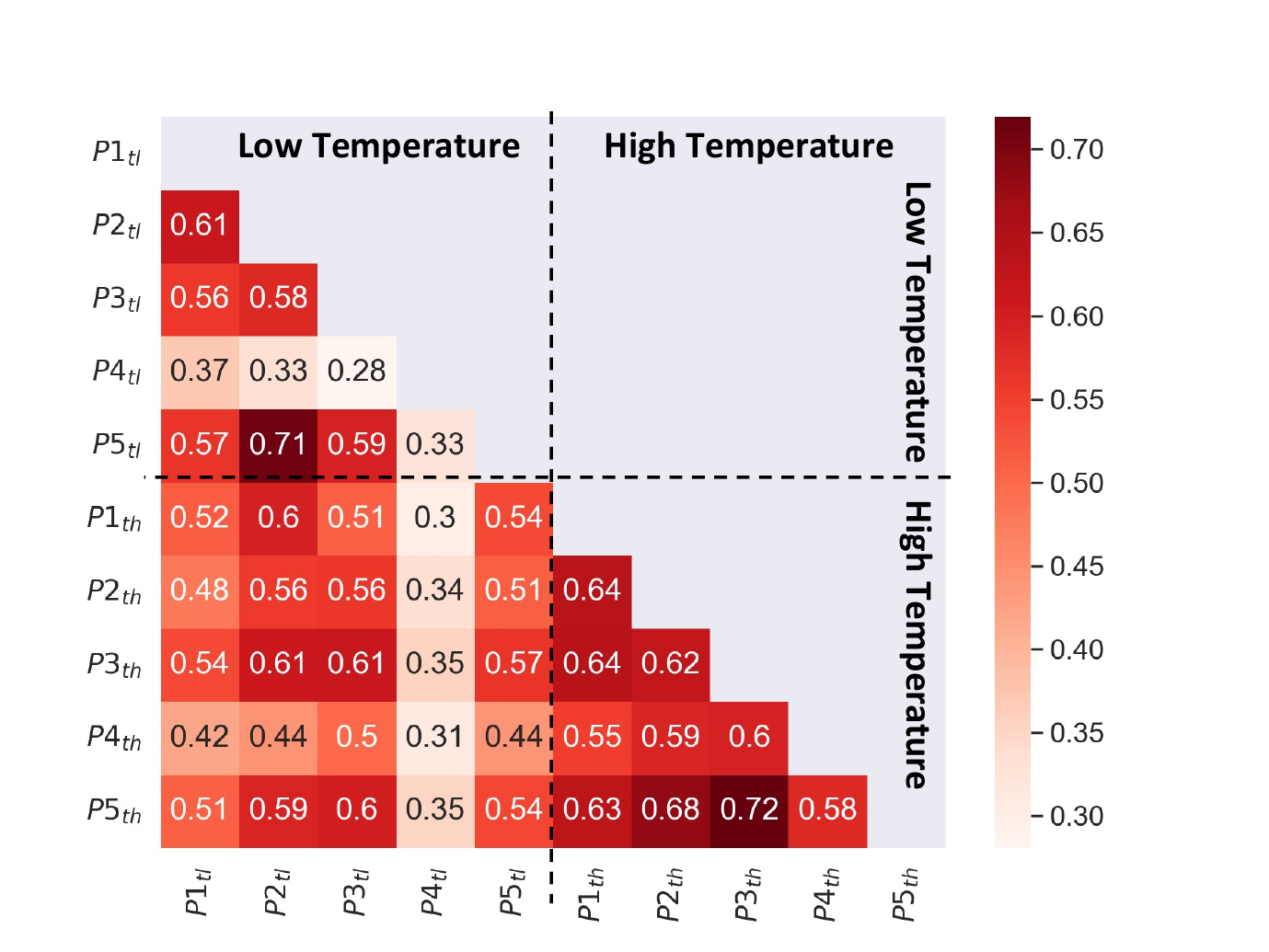}
  \caption{Kendall's Tau correlation between BLEURT scores of various prompts and temperatures in \textsc{no\_src}}
  \label{fig:bwosrc}
\end{figure}

 \begin{table}[H] 
\caption{Comparison of lengths of generated responses by question (Q) vs. statement (S) in the \textsc{wsrc} setting. Question versions of the prompts generate fewer words on average, than their statement counterparts.}
\begin{tabular}{ c| c| c}
Prompt Pair & Avg. Len. (S) & Avg. Len. (Q) \\
\hline
P2-P5 & 43.93 & 34.53 \\
P3-P6 & 32.55 & 31.4 \\
P4-P7 & 42.51 & 32.72
 \label{tab_len_NO_SRC_q}
\end{tabular}
\end{table}
 
\begin{table}[H]
\caption{Comparison of lengths of generated responses by low and high temperatures in the \textsc{no\_src} setting. High temperature generates consistently longer analogies. Same trend is observed in other settings also.}
\begin{tabular}{ c| c| c}
Prompt & Avg. Length (tl) & Avg. Length (th) \\
\hline
P1 & 39.74 & 47.62 \\
P2 & 32.67 & 40.71 \\
P3 & 40.06 & 46.62 \\
P4 & 32.51 & 40.13 \\ 
P5 & 36.53 & 38.50 
 \label{tab_len_wosrc}
\end{tabular}
\end{table}

\section{Mturk study details}
\label{mturk}
\begin{figure*}
  \includegraphics[width=\linewidth]{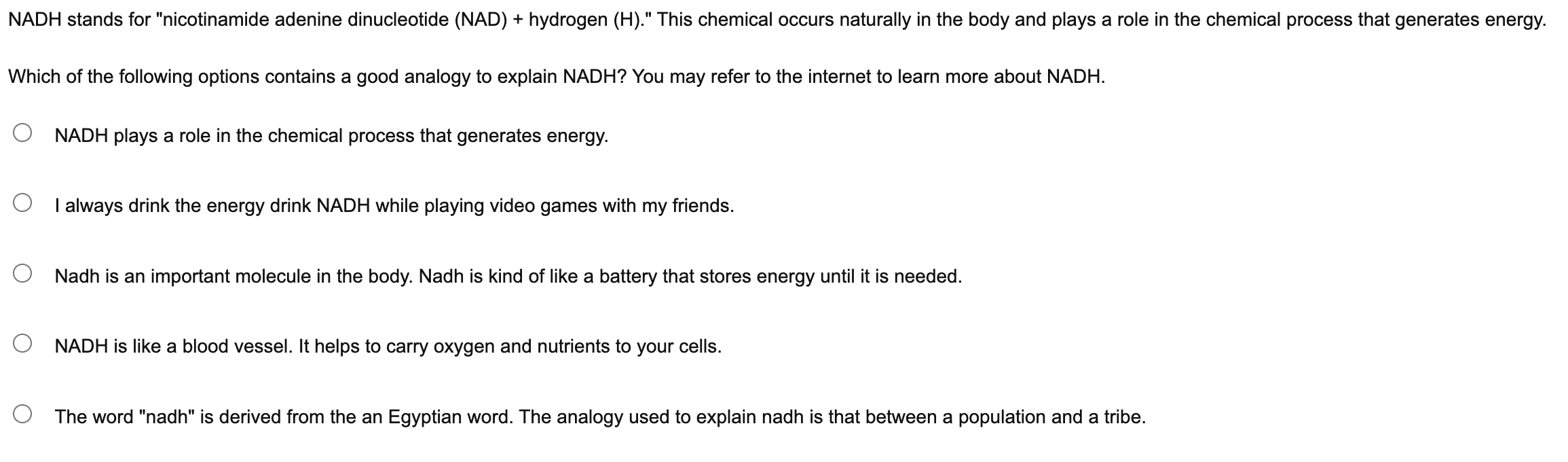}
  \caption{Pre-screening question for identifying qualified workers.}
  \label{fig:int_screen1}
\end{figure*}

For identifying qualified workers on Amazon Mechanical Turk, we designed a pre-screening test (Mturk Qualification) asking them to identify the meaningful analogy for a target concept (Figure \ref{fig:int_screen1}, Appendix \ref{interface}). Further, we used the following additional qualifications: workers should have completed at least 5k tasks with >98\% approval rate and be located in the US since the task requires proficiency in english (this way of filtering is not perfect but there is currently no good way to identify native english speakers via Mturk). We did not collect any other demographic or geographic information about the workers. 

\begin{figure*}
  \includegraphics[width=\linewidth]{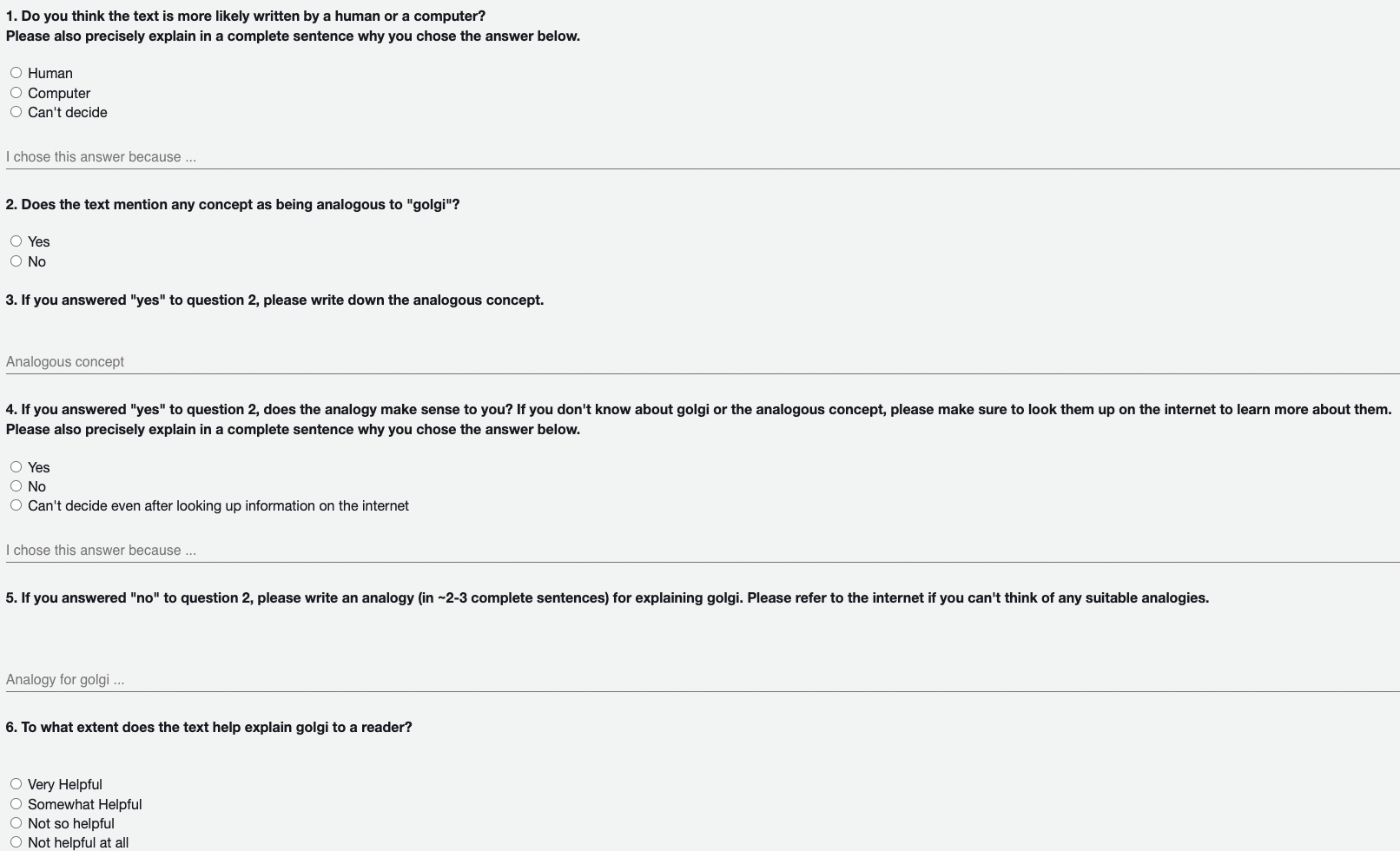}
  \caption{Sample interface for screening qualified workers.}
  \label{fig:int_screen2}
\end{figure*}

Those who passed these qualifications worked on a small test batch of analogies asking detailed questions about their quality (Figure \ref{fig:int_screen2}, Appendix \ref{interface}). The questions consisted of both Likert-style or Binary choice questions and text inputs asking them to explain their choices. We manually assessed their responses, especially paying close attention to their reasoning to identify qualified workers for the main study. 

For both the main study and the screening, a simple definition of the target from sites like Simple English Wikipedia \footnote{https://simple.wikipedia.org/wiki/Main\_Page} was provided to workers as reference and they were encouraged to refer to the internet to learn more about the shown concepts. We also provided several sample annotations as part of the instructions to guide workers. Moreover, we were available to answer clarification questions via a shared chatroom.

Annotators were paid at the rate of \$50/hr. The rate was decided based on open discussions with them and is above the minimum wage. They were informed that the data generated would be used for research purposes. We consulted with our university ethics board and found that IRB was not required for this study.

\section{Human evaluation interface}
\label{interface}

\begin{figure*}
  \includegraphics[width=\linewidth]{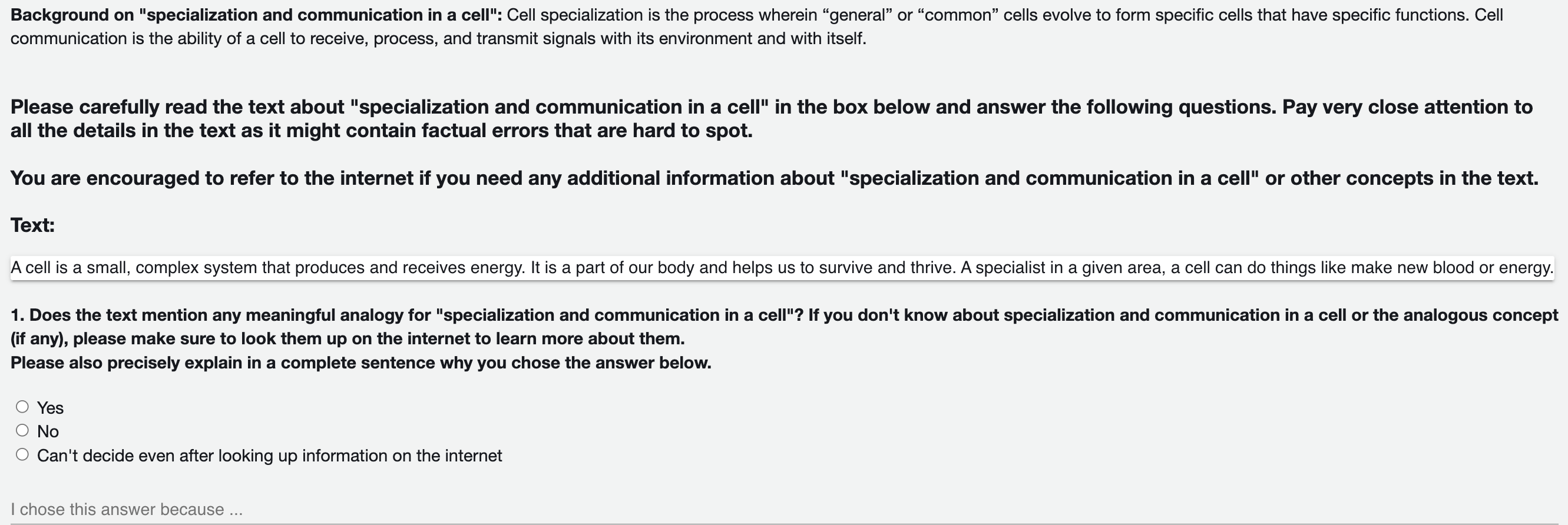}
  \caption{Sample interface for human evaluation of the analogies.}
  \label{fig:int_main}
\end{figure*}

%% file: main.bbl
\begin{thebibliography}{32}
\expandafter\ifx\csname natexlab\endcsname\relax\def\natexlab#1{#1}\fi

\bibitem[{Bender et~al.(2021)Bender, Gebru, McMillan-Major, and
  Shmitchell}]{bender2021dangers}
Emily~M Bender, Timnit Gebru, Angelina McMillan-Major, and Shmargaret
  Shmitchell. 2021.
\newblock On the dangers of stochastic parrots: Can language models be too big?
\newblock In \emph{Proceedings of the 2021 ACM Conference on Fairness,
  Accountability, and Transparency}, pages 610--623.

\bibitem[{Brown et~al.(2020)Brown, Mann, Ryder, Subbiah, Kaplan, Dhariwal,
  Neelakantan, Shyam, Sastry, Askell et~al.}]{brown2020language}
Tom~B Brown, Benjamin Mann, Nick Ryder, Melanie Subbiah, Jared Kaplan, Prafulla
  Dhariwal, Arvind Neelakantan, Pranav Shyam, Girish Sastry, Amanda Askell,
  et~al. 2020.
\newblock Language models are few-shot learners.
\newblock \emph{arXiv preprint arXiv:2005.14165}.

\bibitem[{Callison-Burch et~al.(2006)Callison-Burch, Osborne, and
  Koehn}]{callison2006re}
Chris Callison-Burch, Miles Osborne, and Philipp Koehn. 2006.
\newblock Re-evaluating the role of bleu in machine translation research.
\newblock In \emph{11th Conference of the European Chapter of the Association
  for Computational Linguistics}.

\bibitem[{Devlin et~al.(2018)Devlin, Chang, Lee, and
  Toutanova}]{devlin2018bert}
Jacob Devlin, Ming-Wei Chang, Kenton Lee, and Kristina Toutanova. 2018.
\newblock Bert: Pre-training of deep bidirectional transformers for language
  understanding.
\newblock \emph{arXiv preprint arXiv:1810.04805}.

\bibitem[{Dou et~al.(2021)Dou, Forbes, Koncel-Kedziorski, Smith, and
  Choi}]{dou2021scarecrow}
Yao Dou, Maxwell Forbes, Rik Koncel-Kedziorski, Noah~A Smith, and Yejin Choi.
  2021.
\newblock Scarecrow: A framework for scrutinizing machine text.
\newblock \emph{arXiv preprint arXiv:2107.01294}.

\bibitem[{Fleiss(1971)}]{fleiss1971measuring}
Joseph~L Fleiss. 1971.
\newblock Measuring nominal scale agreement among many raters.
\newblock \emph{Psychological bulletin}, 76(5):378.

\bibitem[{Forbus et~al.(2017)Forbus, Ferguson, Lovett, and
  Gentner}]{forbus2017extending}
Kenneth~D Forbus, Ronald~W Ferguson, Andrew Lovett, and Dedre Gentner. 2017.
\newblock Extending sme to handle large-scale cognitive modeling.
\newblock \emph{Cognitive Science}, 41(5):1152--1201.

\bibitem[{Jayanthi et~al.(2020)Jayanthi, Pruthi, and
  Neubig}]{jayanthi2020neuspell}
Sai~Muralidhar Jayanthi, Danish Pruthi, and Graham Neubig. 2020.
\newblock Neuspell: A neural spelling correction toolkit.
\newblock \emph{arXiv preprint arXiv:2010.11085}.

\bibitem[{Kendall(1938)}]{kendall1938new}
Maurice~G Kendall. 1938.
\newblock A new measure of rank correlation.
\newblock \emph{Biometrika}, 30(1/2):81--93.

\bibitem[{Kittur et~al.(2019)Kittur, Yu, Hope, Chan, Lifshitz-Assaf, Gilon, Ng,
  Kraut, and Shahaf}]{kittur2019scaling}
Aniket Kittur, Lixiu Yu, Tom Hope, Joel Chan, Hila Lifshitz-Assaf, Karni Gilon,
  Felicia Ng, Robert~E Kraut, and Dafna Shahaf. 2019.
\newblock Scaling up analogical innovation with crowds and ai.
\newblock \emph{Proceedings of the National Academy of Sciences},
  116(6):1870--1877.

\bibitem[{Krishna et~al.(2021)Krishna, Roy, and Iyyer}]{krishna2021hurdles}
Kalpesh Krishna, Aurko Roy, and Mohit Iyyer. 2021.
\newblock Hurdles to progress in long-form question answering.
\newblock \emph{arXiv preprint arXiv:2103.06332}.

\bibitem[{Lavie and Agarwal(2007)}]{lavie2007meteor}
Alon Lavie and Abhaya Agarwal. 2007.
\newblock Meteor: An automatic metric for mt evaluation with high levels of
  correlation with human judgments.
\newblock In \emph{Proceedings of the second workshop on statistical machine
  translation}, pages 228--231.

\bibitem[{Li et~al.(2021)Li, Tang, Zhao, and Wen}]{li2021pretrained}
Junyi Li, Tianyi Tang, Wayne~Xin Zhao, and Ji-Rong Wen. 2021.
\newblock Pretrained language models for text generation: A survey.
\newblock \emph{arXiv preprint arXiv:2105.10311}.

\bibitem[{Lin(2004)}]{lin2004rouge}
Chin-Yew Lin. 2004.
\newblock Rouge: A package for automatic evaluation of summaries.
\newblock In \emph{Text summarization branches out}, pages 74--81.

\bibitem[{Liu et~al.(2021)Liu, Yuan, Fu, Jiang, Hayashi, and
  Neubig}]{liu2021pre}
Pengfei Liu, Weizhe Yuan, Jinlan Fu, Zhengbao Jiang, Hiroaki Hayashi, and
  Graham Neubig. 2021.
\newblock Pre-train, prompt, and predict: A systematic survey of prompting
  methods in natural language processing.
\newblock \emph{arXiv preprint arXiv:2107.13586}.

\bibitem[{Lu et~al.(2021)Lu, Bartolo, Moore, Riedel, and
  Stenetorp}]{lu2021fantastically}
Yao Lu, Max Bartolo, Alastair Moore, Sebastian Riedel, and Pontus Stenetorp.
  2021.
\newblock Fantastically ordered prompts and where to find them: Overcoming
  few-shot prompt order sensitivity.
\newblock \emph{arXiv preprint arXiv:2104.08786}.

\bibitem[{Lucy and Bamman(2021)}]{lucy2021gender}
Li~Lucy and David Bamman. 2021.
\newblock Gender and representation bias in gpt-3 generated stories.
\newblock In \emph{Proceedings of the Third Workshop on Narrative
  Understanding}, pages 48--55.

\bibitem[{Ma et~al.(2020)Ma, Cui, Si, Liu, Wang, and Hu}]{ma2020charbert}
Wentao Ma, Yiming Cui, Chenglei Si, Ting Liu, Shijin Wang, and Guoping Hu.
  2020.
\newblock Charbert: character-aware pre-trained language model.
\newblock \emph{arXiv preprint arXiv:2011.01513}.

\bibitem[{Mikolov et~al.(2013)Mikolov, Chen, Corrado, and
  Dean}]{mikolov2013efficient}
Tomas Mikolov, Kai Chen, Greg Corrado, and Jeffrey Dean. 2013.
\newblock Efficient estimation of word representations in vector space.
\newblock \emph{arXiv preprint arXiv:1301.3781}.

\bibitem[{Mitchell(2021)}]{mitchell2021abstraction}
Melanie Mitchell. 2021.
\newblock Abstraction and analogy-making in artificial intelligence.
\newblock \emph{arXiv preprint arXiv:2102.10717}.

\bibitem[{Newby et~al.(1995)Newby, Ertmer, and
  Stepich}]{newby1995instructional}
Timothy~J Newby, Peggy~A Ertmer, and Donald~A Stepich. 1995.
\newblock Instructional analogies and the learning of concepts.
\newblock \emph{Educational Technology Research and Development}, 43(1):5--18.

\bibitem[{Ouyang et~al.(2022)Ouyang, Wu, Jiang, Almeida, Wainwright, Mishkin,
  Zhang, Agarwal, Slama, Ray et~al.}]{ouyang2022training}
Long Ouyang, Jeff Wu, Xu~Jiang, Diogo Almeida, Carroll~L Wainwright, Pamela
  Mishkin, Chong Zhang, Sandhini Agarwal, Katarina Slama, Alex Ray, et~al.
  2022.
\newblock Training language models to follow instructions with human feedback.
\newblock \emph{arXiv preprint arXiv:2203.02155}.

\bibitem[{Pruthi et~al.(2019)Pruthi, Dhingra, and Lipton}]{pruthi2019combating}
Danish Pruthi, Bhuwan Dhingra, and Zachary~C Lipton. 2019.
\newblock Combating adversarial misspellings with robust word recognition.
\newblock \emph{arXiv preprint arXiv:1905.11268}.

\bibitem[{Raffel et~al.(2019)Raffel, Shazeer, Roberts, Lee, Narang, Matena,
  Zhou, Li, and Liu}]{raffel2019exploring}
Colin Raffel, Noam Shazeer, Adam Roberts, Katherine Lee, Sharan Narang, Michael
  Matena, Yanqi Zhou, Wei Li, and Peter~J Liu. 2019.
\newblock Exploring the limits of transfer learning with a unified text-to-text
  transformer.
\newblock \emph{arXiv preprint arXiv:1910.10683}.

\bibitem[{Rossiello et~al.(2019)Rossiello, Gliozzo, Farrell, Fauceglia, and
  Glass}]{rossiello2019learning}
Gaetano Rossiello, Alfio Gliozzo, Robert Farrell, Nicolas~R Fauceglia, and
  Michael Glass. 2019.
\newblock Learning relational representations by analogy using hierarchical
  siamese networks.
\newblock In \emph{Proceedings of the 2019 Conference of the North American
  Chapter of the Association for Computational Linguistics: Human Language
  Technologies, Volume 1 (Long and Short Papers)}, pages 3235--3245.

\bibitem[{Sakaguchi et~al.(2017)Sakaguchi, Duh, Post, and
  Van~Durme}]{sakaguchi2017robsut}
Keisuke Sakaguchi, Kevin Duh, Matt Post, and Benjamin Van~Durme. 2017.
\newblock Robsut wrod reocginiton via semi-character recurrent neural network.
\newblock In \emph{Thirty-first AAAI conference on artificial intelligence}.

\bibitem[{Sellam et~al.(2020)Sellam, Das, and Parikh}]{sellam2020bleurt}
Thibault Sellam, Dipanjan Das, and Ankur~P Parikh. 2020.
\newblock Bleurt: Learning robust metrics for text generation.
\newblock In \emph{Proceedings of ACL}.

\bibitem[{Turney(2008)}]{turney2008latent}
Peter~D Turney. 2008.
\newblock The latent relation mapping engine: Algorithm and experiments.
\newblock \emph{Journal of Artificial Intelligence Research}, 33:615--655.

\bibitem[{Ushio et~al.(2021)Ushio, Espinosa-Anke, Schockaert, and
  Camacho-Collados}]{ushio2021bert}
Asahi Ushio, Luis Espinosa-Anke, Steven Schockaert, and Jose Camacho-Collados.
  2021.
\newblock Bert is to nlp what alexnet is to cv: Can pre-trained language models
  identify analogies?
\newblock \emph{arXiv preprint arXiv:2105.04949}.

\bibitem[{van~der Lee et~al.(2021)van~der Lee, Gatt, van Miltenburg, and
  Krahmer}]{van2021human}
Chris van~der Lee, Albert Gatt, Emiel van Miltenburg, and Emiel Krahmer. 2021.
\newblock Human evaluation of automatically generated text: Current trends and
  best practice guidelines.
\newblock \emph{Computer Speech \& Language}, 67:101151.

\bibitem[{Weidinger et~al.(2021)Weidinger, Mellor, Rauh, Griffin, Uesato,
  Huang, Cheng, Glaese, Balle, Kasirzadeh et~al.}]{weidinger2021ethical}
Laura Weidinger, John Mellor, Maribeth Rauh, Conor Griffin, Jonathan Uesato,
  Po-Sen Huang, Myra Cheng, Mia Glaese, Borja Balle, Atoosa Kasirzadeh, et~al.
  2021.
\newblock Ethical and social risks of harm from language models.
\newblock \emph{arXiv preprint arXiv:2112.04359}.

\bibitem[{Zhao et~al.(2021)Zhao, Wallace, Feng, Klein, and
  Singh}]{zhao2021calibrate}
Tony~Z Zhao, Eric Wallace, Shi Feng, Dan Klein, and Sameer Singh. 2021.
\newblock Calibrate before use: Improving few-shot performance of language
  models.
\newblock \emph{arXiv preprint arXiv:2102.09690}.

\end{thebibliography}
